\documentclass{article}
\usepackage{spconf,amsmath,graphicx}
\usepackage{lipsum}
\usepackage{multirow}
\usepackage{multicol}
\usepackage{svg}
\usepackage{amsmath}
\usepackage{graphicx}
\usepackage{booktabs}
\usepackage{url}

\title{Exploring Feature Representation and Training strategies in Temporal Action Localization}
\name{Tingting Xie$^{\star }$%\thanks{*Work done in part during the internship at CASIA} 
\qquad Xiaoshan Yang$^{\dagger}$ \qquad Tianzhu Zhang$^{\dagger}$ \qquad Changsheng Xu$^{\dagger}$ \qquad Ioannis Patras$^{\star }$}
\address{$^{\star }$ EECS, Queen Mary University of London \qquad
$^{\dagger}$ Institute of Automation, Chinese Academy of Sciences}
\begin{document}
%\ninept
%
\maketitle
\begin{abstract}
Temporal action localization has recently attracted significant interest in the Computer Vision community. However, despite the great progress, it is hard to identify which aspects of the proposed methods contribute most to the increase in localization performance. To address this issue, we conduct ablative experiments on feature extraction methods, fixed-size feature representation methods and training strategies, and report how each influences the overall performance. Based on our findings, we propose a two-stage detector that outperforms the state of the art in THUMOS14, achieving a mAP@tIoU=0.5 equal to $44.20\%$. 
%Our work provides a solid baseline for temporal action localization -- our code will be released to facilitate further research.
%Temporal action localization is an area that has recently attracted significant interest in the Computer Vision community, where several good performing frameworks have been proposed. However, despite the progress, it is hard to identify which aspects of the frameworks contribute most to the increase in the localization performance. To address this issue, we conduct ablative experiments on feature extraction methods, fixed-size representation methods and training strategies, and report how each influences the overall performance. Based on our findings, we propose a two-stage detector that outperforms the state of the art in THUMOS14 achieving a mAP@tIoU=0.5 equal to $44.20\%$. Our work provides a solid baseline for temporal action localization -- our code will be released to facilitate further research.
%
\end{abstract}
\begin{keywords}
Action localization, Temporal structure
\end{keywords}
%--------------------------------------------------------------------------------------------------------
\section{Introduction}
\label{sec:intro}

Temporal action localization, that is recognition of the action class and localization of its temporal boundaries (start and end) in untrimmed videos, has drawn increasing attention from the research community, due to its numerous applications in surveillance, video analysis, video search and other areas \cite{oneata2013action, yeung2016end}. However, despite the significant progress due to the advances in CNN architectures ~\cite{simonyan2014two, tran2015learning, wang2016temporal,carreira2017quo, wang2017untrimmednets}, the performance of action localization methods remains low \cite{zhao2017temporal, gao2017cascaded, chao2018rethinking, lin2018bsn} and methods are not directly comparable as they use different feature extractors, different methods of generating fixed-size features and different training strategies.

More specifically, first, several works adopt different features \cite{tran2015learning, zhao2017cuhk, carreira2017quo}, but it is not clear how much difference feature extractor could make to performance. Second, action localization involves classifying temporal segments of varying length -- sometimes, the variations are significant ranging from a few tens of frames to several hundred frames. Typically, Computer Vision methods generate fixed-size features. Several methods use average temporal pooling \cite{shou2016temporal, shou2017cdc, gao2017turn, gao2017cascaded} over the temporal segment duration. This, eliminates the temporal structure and leads to false positive proposals, for example incomplete actions and multi-action clips. Other methods, preserve the temporal structure, either by using Structured Temporal Pyramid Pooling (STPP) \cite{zhao2017temporal} or linear interpolated sampling \cite{lin2018bsn}. However, STPP may not be the best trade-off between computation and performance, and linear interpolation is not an immediately obvious choice for long video clips as it essentially reduces to a sampling operation. Finally, several methods adopt two stream architectures, such as, \cite{chao2018rethinking} performs early/late fusion on the top of networks built on separate modalities, \cite{gao2017cascaded} refines temporal boundaries in a cascaded way, it is still hard to tell how these strategies have an effect on the final performance.

\begin{figure*}
\begin{center}
% \includesvg[width=\textwidth]{image/cbr.svg}
\includegraphics[width=\textwidth]{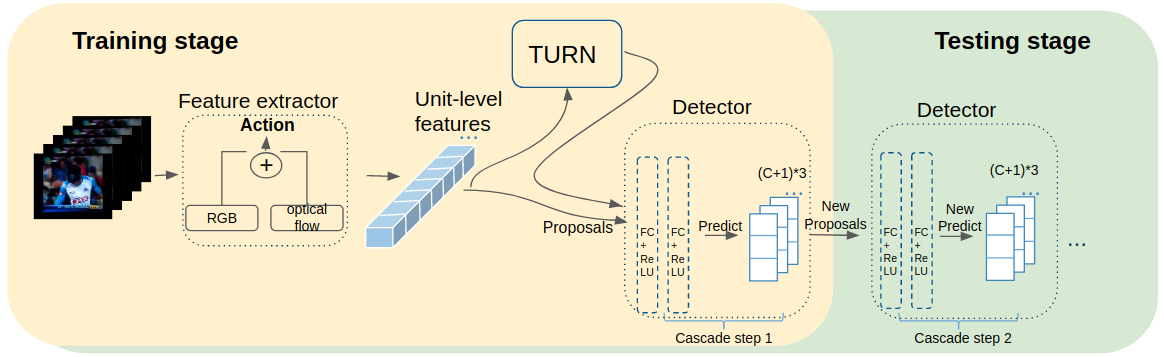}
\end{center}
\caption{Two-stage Cascaded Boundary Regression (CBR) framework. Given an untrimmed video, unit-level features are extracted by feature extractor, in which each unit represents 16 frames. Then, features and sliding window proposal are passed to Temporal Unit Regression Network (TURN)~\cite{gao2017turn} to generate candidate proposals. After that, detectors take proposals as input, and output a confidence score, indicating which class the proposal belongs, and two regression offsets of start and end. In this stage, the temporal boundaries are adjusted in a cascaded way by feeding the refined clips back to the system for further boundary refinement. All the parameters in each cascade step are shared.}
\label{fig:cbr}
\end{figure*}

In this paper, we use as backbone a state of the art, temporal action localization method \cite{gao2017cascaded} (Fig. \ref{fig:cbr}) and investigate the influence of architectural choices with respect to the factors mentioned above. First, we compare the current popular feature extractors, namely \cite{zhao2017cuhk, carreira2017quo}. Second, we investigate into fixed-size feature representation methods, such as average temporal pooling \cite{shou2016temporal, shou2017cdc, gao2017turn, gao2017cascaded}, structured temporal pyramid pooling (STPP) \cite{zhao2017temporal}, linear interpolated sampling \cite{lin2018bsn}. We show that a simple scheme by which we divide the candidate proposals into $k$ parts, do average pooling for each part and concatenate them to be fixed-size feature outperforms the methods above and STPP for equal size representations when $k>=3$. Third, we conduct experiments with different number of cascade steps, and show that several cascade steps improve the performance significantly. Finally, we investigate both on late and early fusion and, contrary to \cite{chao2018rethinking} find that late fusion, and in general appearance information, does not provide in general benefits. By combining our findings, we propose a two-stage method that outperforms the state of the art (mAP@tIoU=0.5 on THUMOS14~\cite{THUMOS14} is $44.20\%$, vs $42.80\%$ by \cite{chao2018rethinking}).
%
% XXXXXXXXXXXXXX Modify it into colors to indicate how you train the model and how you test the model, also add the relu layer and softmax layer
% -------------------------------------------------------------------------------------------------------
\section{Methods}
\label{sec:method}
Our paper builds on a state of the art method, namely Cascaded Boundary Regression (CBR, Fig.~\ref{fig:cbr}), which adopts the classical two-stage framework and, at the first stage performs a Temporal Proposal Generation (TPG)~\cite{gao2017turn}, and in the second stage performs action Detection (DET)~\cite{gao2017cascaded}. TPG and DET networks are both classification/regression networks that take as input a fixed-size feature extracted from temporal clips of varying lengths. The former, takes as input a temporal clip that has been obtained by a sliding window approach (at various temporal durations) and performs binary action/background classification determining whether the input temporal segment is an action or not, and regression, adjusting the start and end of the temporal segment. The latter, performs multi-class classification, taking as input the proposals of the first stage and a) determining the action to which they belong, and b) adjusting the proposal's boundaries. Both stages are the same in terms of the network that implements them so we focus our ablative experiments on DET network part. 

Formally, let us assume that $P$ is a proposal, which is divided into $n$ units, $P=\{u_i\}_1^n$, each of which is represented by a vector $\mathbf{f}_{u_i} \in R^d$. This is common practice \cite{zhao2017cuhk, carreira2017quo} in the field and the feature vector $\mathbf{f}_{u_i}$ is typically obtained by different two-stream networks from action recognition. 

In the following three sections, we investigate three different issues. First, and mainly, different strategies for obtaining a fixed size representation of proposals of varying length (i.e., different size $n$) - we will analyze three popular methods, namely, Pooling~\cite{shou2016temporal,shou2017cdc,gao2017turn,gao2017cascaded}, STPP~\cite{zhao2017temporal} and linear interpolated sampling~\cite{lin2018bsn}. Second, we compare detection performance with different cascade steps (Fig.~\ref{fig:cbr}). Third, we investigate different, namely early vs late, fusion strategies.

%--------------------------------------------------------------------------------------------------------

\subsection{Representations}
\subsubsection{Pooling}
\label{subsec:pooling}
One of the simplest methods to obtain a fixed size representation for a proposal $P$ of arbitrary length is pooling ~\cite{shou2016temporal, shou2017cdc, gao2017turn, gao2017cascaded}. A global average pooling, for example, generates:
\begin{equation}
\label{equ:f_avg}
\mathbf{f}_{P} = \frac{1}{n}\sum_{i \in (1,2,...,n)}{\mathbf{f}_{u_i}}
\end{equation}
Clearly, global pooling disregards any temporal information and the average operation may also mask salient parts of the video -- both, can lead to mis-classifications. Thus, to keep the temporal information, and in contrast to global pooling, we divide it into $k$ parts, without any overlap, do average pooling to each part, and obtain our final representation by concatenating the features of the parts. Similar to \cite{gao2017turn, gao2017cascaded, chao2018rethinking}, we also take context information into consideration, that is consider $n_{ctx}$ units $P_s=\{u_i|i \in (n-n_{ctx},...,n-1)\}$ before, and $n_{ctx}$ units $P_e=\{u_i|i \in (n+1,...,n+n_{ctx})\}$ after the proposal $P$. Clearly, $P_s$ and $P_e$ provide a temporal context for proposals -- this is shown to be important for inferring temporal boundary. Following~\cite{gao2017cascaded}, we set the number of context units to be $2$, and context features are pooled from unit-level features by mean pooling operation. Thus, the final representation of a proposal is the concatenation of context features and the internal feature as shown in Eq.~\ref{equ:f}. % XXX Either say in (2) or say in Eq.2
\begin{equation}
\label{equ:f}
\mathbf{f}_{P}^{Pooling}=\mathbf{f}_{P_s}||{\mathbf{f}_{P_i}}_1^k||\mathbf{f}_{P_e}
\end{equation}
%
%--------------------------------------------------------------------------------------------------------
\subsubsection{Structured Temporal Pyramid Pooling(STPP)}
\label{subsec:stpp}
Another feature representation method is STPP~\cite{zhao2017temporal}, which increases the model's ability to identify different stages of the action.  The proposal $P$ is divided into three stages $P_s$, $P_c$ and $P_e$, representing $starting$, $course$, and $ending$ (the first and the last including context information). The $course$ stage represents the activity process itself, which is constructed by a $\mathcal{L}$-level temporal pyramid where each level evenly divides the interval into $B_l$ parts. Here, we conduct experiments on a two-level (L=2, $B_1=1$, $B_2=2$) and a three-level(L=3, $B_1=1$, $B_2=2$, $B_3=4$) pyramid for the $course$ stage, and simpler one-level pyramids (which essentially reduces to standard average pooling) for $starting$ and $ending$ parts. Finally, the stage-wise features are combined via concatenation. This construction explicitly has a better understanding of global and local content, but it takes lots of computation resource while expressing fine-grained temporal structure.
% (which means divide $course$ into 2 parts and concatenate all the global and local feature)
% --------------------------------------------------------------- -----------------------------------------
\subsubsection{Linear interpolated sampling}
\label{subsec:sampling}
Finally, \cite{lin2018bsn} extracts a fixed length feature, called Boundary-Sensitive Proposal (BSP) feature, by sampling. Given a candidate proposal, BSP samples the sequence (feature vector sequence) to get $\mathbf{f}_P$ by linear interpolation with 16 points. In starting and ending context regions, BSP samples with 8 linear interpolation points and gets $\mathbf{f}_{P_s}$ and $\mathbf{f}_{P_e}$ separately. Concatenating these vectors, results to the final feature $\mathbf{f}_P^{BSP}=({\mathbf{f}_{P_s}}_1^A || {\mathbf{f}_{P}}_1^B || {\mathbf{f}_{P_e}}_1^A)$ where $A/B/A = 8/16/8$ are the number of interpolated points used in \cite{lin2018bsn} -- here we also report results with $2/4/2$ and $4/8/4$ that performed better.

%---------------------------------------------------------------------------------------------------------
%
\subsection{Cascade} 
Several recent methods stack detection stages one after the other and significantly improve a baseline detector. Following \cite{gao2017cascaded} we adopt a three step cascade to adjust the temporal boundaries by feeding the refined clips back to the system for further boundary refinement.

\subsection{Early vs late fusion.}
We investigate both dominant paradigm for fusing motion and texture: early fusion (by concatenating input features), and late fusion (by averaging the results from two single-stream learners). Contrary to \cite{chao2018rethinking} we find that fusion does not contribute much -- and in particular that texture information is much inferior to motion in this context and method.

%early fusion is preferable than late fusion. More specifically, we find that appearance information does not help in a late fusion scheme, but early fusion improves the performance in comparison to a single modality.
%There are two dominant ways of fusion of multiple sources of information (motion and appearance in our case): early fusion (fusion by concatenating features), and late fusion (fusion by averaging/weighing the results from two single-stream learners). Here, we investigate both and, contrary to \cite{chao2018rethinking} we find that early fusion is preferable than late fusion. More specifically, we find that appearance information does not help in a late fusion scheme, but early fusion improves the performance in comparison to a single modality.

%---------------------------------------------------------------------------------------------------------
\section{Implementations/Dataset/Evaluation}
\label{sec:dataset}
\textbf{Implementation details.} Each step of the detection network in CBR comprises of two fully-connected (fc) layers mainly, the input dimension of the first fc is $n_f$, which varies depending on the feature we use, and the output dimension is $1000$. For the second layer, the output dimension is $(C+1)*3$ where $C$ is the number of classes. During training, we use a batch size of $128$, an original learning rate of $0.001$ for two-stream training, and $0.005$ for each stream training -- we will report the updating strategy in the corresponding sections. All the reported results are the mean of three experiments, where the networks we trained for $50000$ iterations. Unless stated otherwise we use the candidate proposals from~\cite{gao2017turn}. 

\textbf{Dataset.} THUMOS14~\cite{THUMOS14} dataset contains 200 and 213 temporal annotated untrimmed videos with 20 action classes in validation and testing set separately. Since there is no training dataset for it (UCF101~\cite{UCF101} is used instead), following the standard practice~\cite{zhao2017temporal, gao2017cascaded, chao2018rethinking}, we train our models on the validation set and evaluate them on the testing set.
%THUMOS14~\cite{THUMOS14} dataset contains 200 and 213 temporal annotated untrimmed videos with 20 action classes in validation and testing set separately. This dataset doesn't provide the training set by itself. Instead, the UCF101~\cite{UCF101}, a trimmed video dataset is used as the training set. Following the standard practice~\cite{zhao2017temporal, gao2017cascaded, chao2018rethinking}, we train our models on the validation set and evaluate them on the testing set. 

\textbf{Evaluation metrics.} We report the mean Average Precision (mAP), where the Average Precision(AP) is calculated for each action class respectively. To compare with others on THUMOS14, we report the mAP with tIoU (temporal Intersection over Union) thresholds at $\{0.3,0.4,0.5,0.6,0.7\}$.
%In temporal action localization task, we report the mean Average Precision (mAP), where the Average Precision(AP) is calculated for each action class respectively. On THUMOS14, we report the mAP with tIoU (temporal Intersection over Union) thresholds $\{0.3,0.4,0.5,0.6,0.7\}$.
% In temporal proposal generation task, we report the Average Recall(AR) calculated with multiple tIoU (temporal Intersection over Union) thresholds. Following conventions, we use tIoU thresholds set $[0.5:0.05:1.0]$ in THUMOS14. To evaluate the relation between recall and the number of proposals, we report the AR at several Average Number of Proposals (AN) on both datasets (denoted as AR@AN). 
 
%---------------------------------------------------------------------------------------------------------
\section{Experimental results}
\label{sec:as}
\subsection{Feature extractor}
We compare the localization performance of two feature extractors: anet-cuhk \cite{zhao2017cuhk} and I3D~\cite{carreira2017quo}. Both models take a stack of 16 RGB/optical flow frames as input, perform spatio-temporal convolutions, and extract a 2048 (anet-cuhk)/1024 (I3D)-dimension feature as the output of an average pooling layer. To compute optical flow, we use DenseFlow \cite{TSN2016ECCV}, as in \cite{zhao2017temporal}. Thus, the input to our action localization model is two 2048/1024-dimension feature maps (for RGB and optical flow) divided stack by stack -- for each stack, we concatenate the RGB feature and optical flow feature. As stated in section~\ref{subsec:pooling}, we fix the pooling level $k$ to $5$, and use the cascade settings in \cite{gao2017cascaded}, and report the localization result mAP@tIoU in Table \ref{tab:feat_extractor}. Our experiments demonstrate that the two features have similar performance -- as the anet-cuhk \cite{zhao2017cuhk} typically performs better at $tIoU=0.5$, we use this in all of our subsequent experiments.
\begin{table}
\vspace*{-5mm}
\begin{center}
\caption{mAP@tIoU=0.5(\%) with different feature extractor.} 
\label{tab:feat_extractor}
\begin{tabular}{cccccccc} \toprule
{tIoU} & {0.3} & {0.4} & {0.5} & {0.6} & {0.7} \\ \midrule
anet16-cuhk &  57.37 & 52.87 & \textbf{39.00} & 18.65 & \textbf{5.16}
 \\
I3D &  \textbf{59.56} & \textbf{54.45} & 38.73 & \textbf{19.31} & 4.95
 \\\bottomrule
\end{tabular}
\end{center}
\vspace*{-5mm}
\end{table}
%
%Experiment shows that, both feature extractors have almost the same affect on final detection performance with the same setting. After massive experiments, we found that anet-cuhk always performs better at $tIoU=0.5$, so in all the next experiments, we take~\cite{zhao2017cuhk} as the default feature extractor.
%---------------------------------------------------------------------------------------------------------
\subsection{Fixed-size representation methods} 
\begin{figure}[htb]
\begin{center}
\includegraphics[width=7cm]{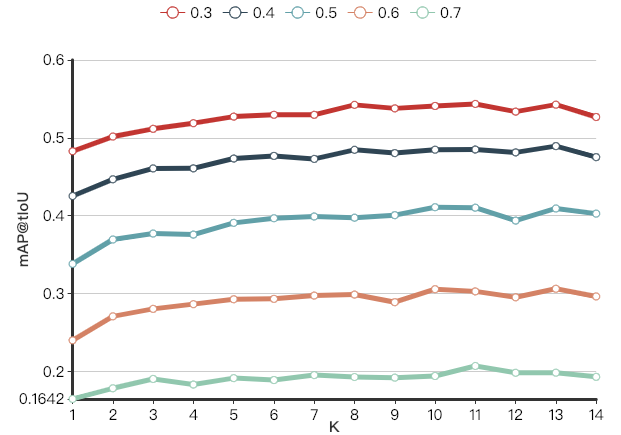}
\end{center}
\caption{mAP@tIoU=0.5 with various $k$. }
\label{fig:pooling}
\end{figure}
To validate how $k$ affects the localization performance, we report the localization performance for different values of $k$ in Fig.\ref{fig:pooling}. We can see that the performance with $k$ increases, but after $k=11$, it starts decreasing. This is something one would expect as higher $k$ maintain better the temporal structure, but also increase the risk of over-fitting. To reduce the computations, in what follows we use $k=5$. 
% 
% %
\begin{table}
\vspace*{-5mm}
\begin{center}
\caption{mAP@tIoU(\%) with different fixed-size feature representation methods.} 
\label{tab:representation}
\begin{tabular}{cccccc} \toprule
{tIoU} & {0.3} & {0.4} & {0.5} & {0.6} & {0.7}  \\ \midrule
STPP(L=2) & 51.42 & 45.20 & 37.23 & 27.04 & 17.92 \\
STPP(L=3) & 51.95 & 46.08 & 38.01 & 27.21 & 18.24 \\
BSP(2/4/2) & 50.96 & 45.86 & 37.25 & 27.51 & 18.15 \\
BSP(4/8/4) & 51.42 & 45.34 & 37.72 & 27.08 & 17.76 \\
BSP(8/16/8) & 49.73 & 43.54 & 35.41 & 25.86 & 17.17 \\
Ours(k=3) & 51.15 & 46.06 & 37.69 & 28.00 & 18.99\\ 
Ours(k=5) & 52.72 & 47.33 & 39.06 & 29.24 & 19.10 \\
Ours(k=10) & \textbf{54.09} & \textbf{48.47} & \textbf{41.07} & \textbf{30.53} & \textbf{19.37}\\\bottomrule
\end{tabular}
\end{center}
\vspace*{-5mm}
\end{table}

\begin{table}
\vspace*{-1mm}
\small\addtolength{\tabcolsep}{-4pt}
\begin{center}
\caption{Number of parameters(M).} 
\label{tab:time}
\begin{tabular}{c|c|c|c|c|c|c|c|c}
\toprule
\multirow{2}{*}{Method} & \multicolumn{2}{c|}{STPP} & \multicolumn{3}{c|}{BSP} & \multicolumn{3}{c}{Ours} \\
\cline{2-9}
& {L=2} & {L=3} & {2/4/2} & {4/8/4} & {8/16/8} & {k=3} & {k=5} & {k=10} \\ \midrule
{\#Params} & 20.54 & 36.93 & 41.02 & 73.79& 139.33 & 20.54 & 28.74 & 49.22\\\bottomrule
\end{tabular}
\end{center}
\vspace*{-5mm}
\end{table}
In Table~\ref{tab:representation} we compare three fixed-size feature representation methods. It is clear that a simple part-divided temporal pooling consistently outperforms STPP and BSP, for equally sized representations.

The only difference between these methods is the number of the parameters used, that itself depends on the input features. % The testing time depends on the number of the parameters with early fusion feature of anet-cuhk~\cite{zhao2017cuhk}, which 
Those are shown in Table \ref{tab:time}. It is clear that, our method uses fewer parameters in comparison to methods of comparable performance.

In Table \ref{tab:proposal_gen} we report the performance with respect to the number of proposals. The performance peaks at $AN=500$ and then slightly drops. In the following section, we take the proposals generated from $AN=500$.
\begin{table}
\vspace*{-5mm}
\begin{center}
\caption{mAP@tIoU=0.5(\%) for different Average Number(AN) proposals. (k=5)} 
\label{tab:proposal_gen}
\begin{tabular}{cccccc} \toprule
{mAP@AN} & {50} & {100} & {300} & {500} & {600} \\ \midrule
TURN & 25.29 & 32.85 & 39.10 & 40.25 & 39.86\\\bottomrule
\end{tabular}
\end{center}
\vspace*{-5mm}
\end{table}
%
%However, we feel that there is a lot of space to investigate into simple and elegant fixed-size feature representation methods for leveraging the temporal structure while maintaining low computation cost.
%---------------------------------------------------------------------------------------------------------
\subsection{Cascade}
In Table \ref{tab:cascade}, we report the performance of using cascade step equals 1 and 3 separately as in \cite{gao2017cascaded}. It is clear that three-step cascade suppresses one-step cascade with more than $2\%$ improvement at different tIoUs.

\begin{table}
\vspace*{-5mm}
\begin{center}
\caption{mAP@tIoU (\%) with different cascade methods.} 
\label{tab:cascade}
\begin{tabular}{cccccc} \toprule
{tIoU} & {0.3} & {0.4} & {0.5} & {0.6} & {0.7} \\ \midrule
cas\_step=1 & 52.01 & 47.10 & 40.25 & 28.81 & 18.93\\
cas\_step=3 &  \textbf{53.45} & \textbf{50.19} & \textbf{44.20} & \textbf{33.93} & \textbf{22.71}\\
 \bottomrule
\end{tabular}
\end{center}
\vspace*{-5mm}
\end{table}
\subsection{Early fusion vs Late fusion}

Table \ref{tab:fusion} reports the results of the two single-stream networks and the early and late fusion schemes with different tIoU. We can draw two conclusions: first, the optical flow feature suppresses the RGB feature -- this is consistent with results reported on action recognition~\cite{simonyan2014two}. Second, that appearance information does not seem to bring significant or consistent benefits in this domain and dataset. Fusion of appearance and motion information marginally outperforms the optical flow only network and only at tIoU=0.5 -- in all other cases, motion information above achieves the best results. This seems to contradict what is reported in ~\cite{chao2018rethinking}. %We can also see in Fig.\ref{fig:fusion}, that the model of the RGB feature converges quite fast at the beginning, and after $10000$ iterations, the model is over-fitting. This may affect negatively the performance of early fusion and late fusion. 
\begin{table}
% \vspace*{-5mm}
\begin{center}
\caption{mAP@tIoU (\%) with different fusion methods.} 
\label{tab:fusion}
\begin{tabular}{cccccc} \toprule
{tIoU} & {0.3} & {0.4} & {0.5} & {0.6} & {0.7} \\ \midrule
RGB &  42.56 & 36.13 & 29.81 & 21.42 & 14.45 \\
Flow &  \textbf{54.38} & \textbf{50.33} & 43.93 & \textbf{34.25} & \textbf{24.57} \\
Early fusion&  53.45 & 50.19 & \textbf{44.20} & 33.93 & 22.71 \\
Late fusion &  52.41 & 47.52 & 40.11 & 29.52 & 18.41 \\
 \bottomrule
\end{tabular}	
\end{center}
\vspace*{-5mm}
\end{table}
%
%\begin{figure}[htb]
%\begin{center}
%\includegraphics[width=7cm]{image/fusion.png}
%\end{center}
%\caption{mAP@tIoU=0.5 (\%) with different iteration steps. }
%\label{fig:fusion}
%\end{figure}
%
\subsection{Comparison with state-of-the-art.}
Finally, we compare the proposed method with the state-of-the-art in Table \ref{tab:comp} and show that we consistently outperform them, for $k=5$ for which we performed most of the ablation studies, for $k=11$ for which we obtained our best results and for all values in between.
\begin{table}
\vspace*{-5mm}
\begin{center}
\caption{Action localization mAP@tIoU (\%).} 
\label{tab:comp}
\begin{tabular}{cccccc} \toprule
{tIoU} & {0.3} & {0.4} & {0.5} & {0.6} & {0.7} \\ \midrule
Gao et al.~\cite{gao2017cascaded} & 50.1 & 41.3 & 31.0 & 19.1 & 9.9 \\
Zhao et al.~\cite{zhao2017temporal} & 51.9 & 41.0 & 29.8 & - & - \\
Lin et al.~\cite{lin2018bsn} & 53.5 & 45.0 & 36.9 & 28.4 & 20.0 \\
Chao et al.~\cite{chao2018rethinking} & 53.2 & 48.5 & 42.8 & 33.8 & 20.8 \\ \midrule
Ours &  \textbf{53.5} & \textbf{50.2} & \textbf{44.2} & \textbf{33.9} & \textbf{22.7} \\
 \bottomrule
\end{tabular}	
\end{center}
\vspace*{-5mm}
\end{table}
%---------------------------------------------------------------------------------------------------------
\section{Conclusion}
\label{conclusion}
In this work we investigate several design choices and their influence in temporal action localization. We propose a simple representation that maintains the temporal structure, and a scheme that calculates the weighted averages over the decisions at several steps of a cascade. We show that our proposed variations lead to large improvements in comparison to the baseline method and in comparison to the State of the Art.

% --------------------------------------------------------------------------------------------------------
%\section{Acknowledgements}
%
\vspace*{2ex}
\noindent {\bf ACKNOWLEDGMENTS}
The work of Tingting Xie is supported by the China Scholarship Council (CSC) and QMUL. This work has been supported by the Royal Society Newton Mobility Grant. We would like to thank NVIDIA for the donation of GPU cards.
%
% --------------------------------------------------------------------------------------------------------
\bibliographystyle{IEEEbib}
\bibliography{icip19}

\end{document}